\pdfoutput=1

\documentclass[11pt]{article}

\usepackage[preprint]{acl}

\usepackage{times}
\usepackage{latexsym}

\usepackage[T1]{fontenc}

\usepackage[utf8]{inputenc}

\usepackage{microtype}

\usepackage{inconsolata}

\usepackage{graphicx}
\usepackage{subcaption}
\usepackage{multirow}
\usepackage{comment}
\usepackage{longtable}
\usepackage{amsmath}

\newcommand{\jmf}[1]{} 
\newcommand{\cm}[1]{} 

\newcommand{\model}[1]{#1}

\author{Changmao Li \\
   University of California, Santa Cruz \\
  \texttt{changmao.li@ucsc.edu} \\\And
  Jeffrey Flanigan \\
  University of California, Santa Cruz \\
  \texttt{jmflanig@ucsc.edu} \\}

\title{Efficient and Effective LLM Factuality Post-Correction with Retrieval}

\title{RAC: Efficient LLM Factuality Correction with Retrieval Augmentation}

\begin{document}
\maketitle
\begin{abstract}
Large Language Models (LLMs) exhibit impressive results across a wide range of natural language processing (NLP) tasks, yet they can often produce factually incorrect outputs. This paper introduces a simple but effective low-latency post-correction method, \textbf{Retrieval Augmented Correction (RAC)}, aimed at enhancing the factual performance of LLMs without requiring additional fine-tuning. Our method is general and can be used with any instruction-tuned LLM, and has greatly reduced latency compared to prior approaches. RAC decomposes the LLM's output into atomic facts and applies a fine-grained verification and correction process with retrieved content to verify and correct the LLM-generated output. Our extensive experiments show that RAC yields up to 30\% improvements over state-of-the-art baselines across two popular factuality evaluation datasets, validating its efficacy and robustness in both with and without the integration of Retrieval-Augmented Generation (RAG) across different LLMs.\footnote{Our code is at \url{https://github.com/jlab-nlp/Retrieval-Augmented-Correction}}
\end{abstract}

\section{Introduction}
Recently Large Language Models (LLMs) have markedly changed the world of natural language processing \cite{brown2020language, chowdhery2022palm, touvron2023Llama, chatgpt, openai2023gpt4}. Although LLMs can achieve superior performance on many NLP tasks, hallucination is a known issue for LLMs~\cite{rawte2023survey, 10.1145/3571730, zhang2023sirens, ye2023cognitive, huang2023survey}. In particular, factually incorrect content generated by LLMs can be explicitly harmful to the application of LLMs \cite{li2024dawn}, including providing incorrect medical suggestions or wrong information for educational purposes.  Misinformation can cause unpredictable harm to humans when LLMs are broadly used. Enhancing LLMs with better factuality can improve LLMs performance \cite{NEURIPS2022_df438caa} and be less harmful to users.

To alleviate this factuality problem, previous research has investigated incorporating retrieved knowledge from a collection of documents into the LLM's context.  This technique is called retrieval augmented generation (RAG) \cite{chen-etal-2017-reading, guu2020realm,NEURIPS2020_6b493230,izacard2022atlas}. RAG first retrieves from a document set to acquire information related to the input task. Retrieval can be done with a search engine such as Google or a from a corpus such as Wikipedia. The retrieved information is then input to the LLM with the task instructions. This predisposes the LLM to generate content that is faithful to the retrieved content, achieving improved factual performance \cite{NEURIPS2020_6b493230, asai2024reliable}.\jmf{this is a strong claim.  Do we have a citation for this?} \cm{An example citation cited before \cite{NEURIPS2020_6b493230}} However, RAG by itself does not guarantee factual content \cite{wu2024faithful}; even with entirely correct retrieved content in the context, the LLMs can still generate factually incorrect output. One of the reasons that LLMs may still generate incorrect output is due to constraints and uncertainty in their internal states \cite{neeman2022disentqa, mallen-etal-2023-trust}.

\jmf{adding} Prior work has attempted to improve RAG by improving the quality of retrieval \cite{asai2024selfrag} or attempting to correct retrieved content \cite{yan2024corrective}.  Surprisingly, we find that these steps are not necessary, and RAG with Google search\jmf{with very simple post-processing. move into main paper, add ablations on those tricks} produces results that are over ten points higher than previous baselines.  Therefore, we focus on correcting generated output using the retrieved content.  While we are not the first to use retrieved content to improve factuality \cite{gao-etal-2023-rarr}, we greatly improve upon it (see Table~\ref{tab:TruthfulQAresults}).

We propose a method we call \textbf{Retrieval Augmented Correction (RAC)}. 
RAC verifies and corrects LLM-generated content using retrieved knowledge to ensure factuality.  Specifically, RAC breaks down LLM-generated content into atomic facts, leverages retrieved knowledge to verify or correct these atomic facts, and then revises the LLM output accordingly.

Lightweight and general, our approach can be viewed as a post-processing component of RAG to improve factuality performance further. Experiments show that our approach provides substantial improvements over all prior results across two popular factuality evaluation datasets, with improvements up to 30\%.

By retrieving once and correcting once, our approach has greatly reduced latency compared to similar prior methods.  Additionally, our approach demonstrates that in some cases, the performance of our method without RAG can surpass that with RAG, indicating the robustness and effectiveness our method even in the absence of retrieval augmented generation.

In summary, our contributions are the following:
\begin{itemize}
    \item We proposed a plug-and-play post-processing component to RAG which improves the factuality performance of RAG-based LLMs.
    \item The proposed verification, correction, and revision modules are fine-tuning free and can be applied to real applications with RAG or without RAG.
    \item Experimental results show the approach can improve factualty by up to $30\%$, depending on the application, without additional training.  We report experiments on two popular factuality evaluation datasets with four LLMs with and without RAG, verifying the applicability of the approach.
    \item Our approach exhibits greatly reduced latency while achieving similar or better performance compared to prior correction by retrieval methods (Table~\ref{tab:real-latency}).
\end{itemize}

\begin{table*}[t!]
\centering
\begin{tabular}{l|c|c|c|c}
 &
  \multicolumn{1}{l|}{\begin{tabular}[c]{@{}l@{}}\# of generation \\ API calls\end{tabular}} &
  \multicolumn{1}{l|}{\begin{tabular}[c]{@{}l@{}}\# of search \\ queries for \\ each sentence \\ or iteration\end{tabular}} &
  \multicolumn{1}{l|}{\begin{tabular}[c]{@{}l@{}}\# of correction \\ iterations\end{tabular}} &
  \multicolumn{1}{l}{\begin{tabular}[c]{@{}l@{}}total \# \\ of retrieval calls\end{tabular}}  \\ \hline
RARR \cite{gao-etal-2023-rarr}      & 1    & $n_q$ & 1 & $n_s*n_q$   \\
CRITIC \cite{gou2024critic}    & 1    & 1    & 3 & 3       \\
EVER  \cite{kang2024evermitigatinghallucinationlarge}     & $n_s$ & 3    & 2 & 3*$n_s$ \\
RAC (ours) & \textbf{1}    & \textbf{1}   & \textbf{1} & \textbf{1}
\end{tabular}
\caption{Number of API calls and retriever calls compared to previous post-correction with retrieval methods. $n_s$ is the number of generated sentences, and $n_q$ is the number of generated questions per sentence (for RARR only). For experiments measuring latency, see \S\ref{ssec:latency}.}
\label{tab:latency}
\end{table*}

\section{Related Work}

Hallucination has been a known issue for generation tasks, especially when using LLMs \cite{maynez-etal-2020-faithfulness, rawte2023survey, 10.1145/3571730, zhang2023sirens, ye2023cognitive, huang2023survey}. Our work focuses on one of the hallucination types for LLMs, factual incorrectness. There are four lines of work regarding reducing factual incorrectness: 1) from the LLM decoding perspective, 2) from the factual enhancement perspective using retrieval augmentation or fine-tuning, 3) from a self-correction or self-alignment perspective, and 4) from a post-correction using retrieved content perspective.

From the LLM decoding perspective,  \citet{li2023inferencetime} proposes Inference-Time Intervention (ITI) to enhance the truthfulness of large language models (LLMs). ITI shifts model activations during inference, following a learned set of directions across a limited number of attention heads. \citet{chuang2024dola} introduces a new decoding method that contrasts predictions made by different model layers to improve factuality performance. \citet{das2024entropy} proposed extrapolating critical token probabilities beyond the last layer for more accurate contrasting during LLMs decoding. 

From the factual enhancement perspective, there are two sub-types. One sub-type is factuality enhancement training or fine-tuning. \citet{NEURIPS2022_df438caa} proposed a modified top-k sampling strategy and a factuality-enhanced training method to improve the factuality of text generation. \citet{yang2023alignment} focus on honesty-based fine-tuning, empowering LLMs to admit limitations by acknowledging “I don’t know.” \citet{tian2024finetuning} constructed a direct preference optimization (DPO) dataset to fine-tune the LLM to improve factuality using reference-based and reference-free truthfulness annotation techniques.

For the factual enhancement perspective, the second sub-type is Retrieval Augmented Generation (RAG) \cite{chen-etal-2017-reading,NEURIPS2020_6b493230, izacard2022atlas}. Self-RAG \cite{asai2023selfrag} is proposed to selectively retrieve knowledge and introduce a model to decide whether to retrieve it. \citet{yoran2024making} designed an NLI model to identify and remove irrelevant context in RAG and improve robustness. SAIL \cite{luo-etal-2023-search} is tuned on instructions to insert retrieved documents before instructions. \citet{jiang-etal-2023-active} actively anticipate future content and decide when and what to retrieve in long-form generation. \citet{yan2024corrective} designed a lightweight retrieval evaluator to assess the overall quality of retrieved documents for a query to filter out or correct incorrect or irrelevant retrieval content for RAG. 

From the self-correction or self-alignment perspective, \citet{zhang2024selfalignment} proposed a self-evaluation component. They prompt an LLM to validate the factuality of its own generated responses based on its internal knowledge and utilize Self-Knowledge Tuning to augment the LLM's self-evaluation ability by improving the model's confidence estimation and calibration. \citet{wang2024finegrained} proposed a self-endorsement framework that leverages the fine-grained fact-level comparisons across multiple sampled responses. 

Recently, some work has investigated post-correction using retrieved content, the area we focus on. \citet{gao-etal-2023-rarr} introduces RARR, which generates several questions for each output, retrieves Bing Search for each question one by one, and repeatedly revises the output based on the retrieved content by iterating questions. Two major drawbacks to this method are that the retrieval process is expensive due to many calls to the retriever, and the correction for this method is cascaded, which can introduce errors. Our approach instead works by breaking down LLM outputs into atomic facts, verifying these facts against retrieved relevant documents, and then revising the output accordingly. \citet{gou2024critic} introduces CRITC, which iteratively conducts correction using LLMs with retrieved knowledge, where for each iteration, they apply the LLM to generate a query to retrieve the knowledge base and revise the output based on the history of revising and decide the most possible answer. \citet{kang2024evermitigatinghallucinationlarge} introduces EVER, which conducts sentence-by-sentence generation and correction in real-time generation, and for each sentence correction, they retrieve using different features multiple times. Their approach largely increases the latency and possibly introduces more hallucinations. Compared to previous approaches, our approach conducts the correction in a more fine-grained manner all at once, reduces the burden of generating questions and correcting several iterations, avoids more hallucinations by not correcting already corrected atomic facts, and only needs to retrieve the knowledge base once for each output based on their task instructions. Our method improves both efficiency and effectiveness compared to the previous approaches. Table \ref{tab:latency} shows the number of API calls and retriever calls for previous correction methods with retrieved content.

\section{Retrieval Augmented Correction (RAC)}
Our approach to improving factuality is to retrieve documents relevant to the input, and use these documents to revise the output so that it is factual.


\begin{figure*}[t!]
\centering
\includegraphics[scale=0.33]{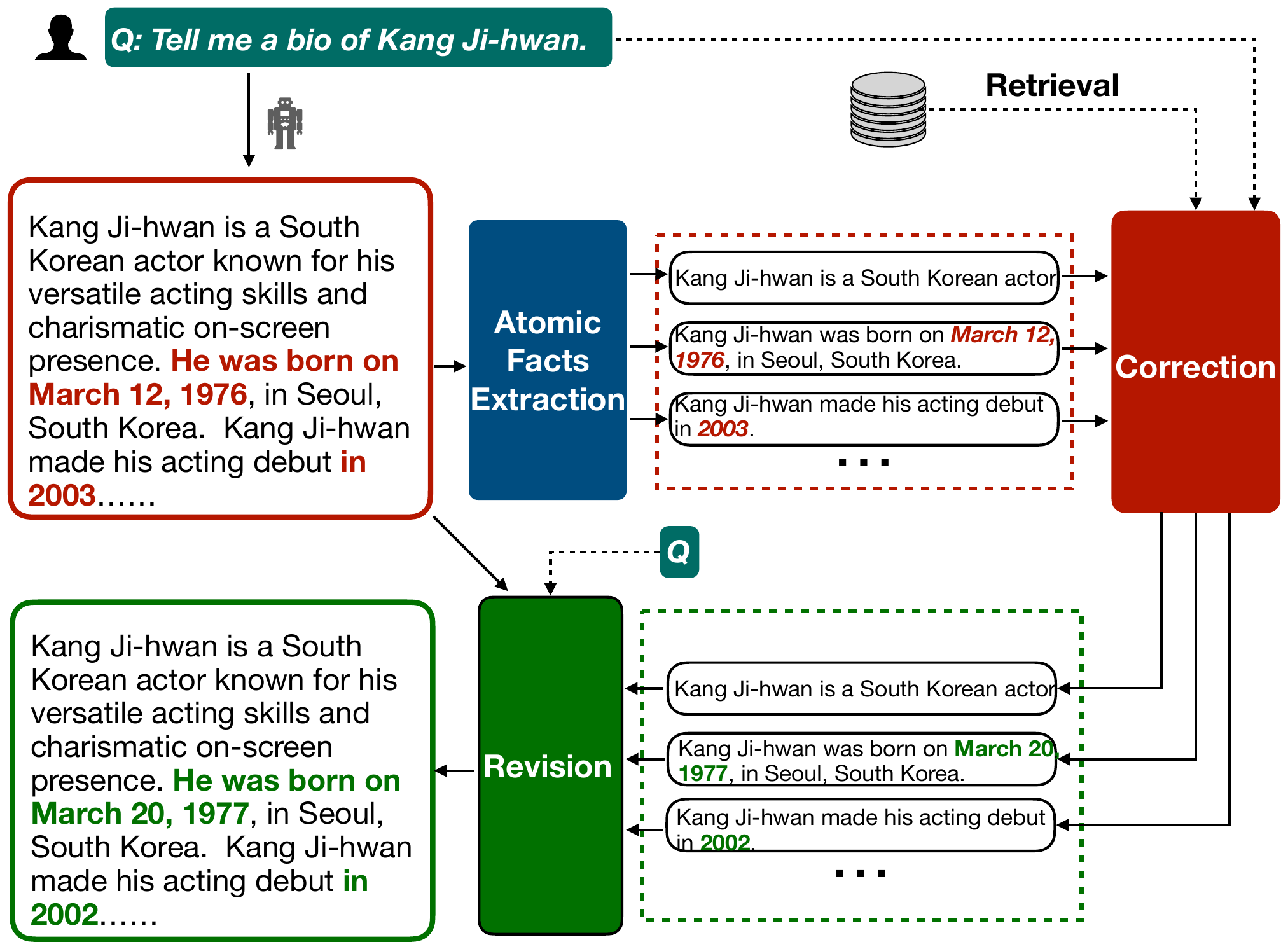}
\caption{Approach overview without using RAG. Note we do not use a verification stage (see Figure~\ref{fig:overview2} below) when not using RAG, since we find that many sentences need to be corrected anyways.}
\label{fig:overview}
\end{figure*}

\begin{figure*}[t!]
\centering
\includegraphics[scale=0.33]{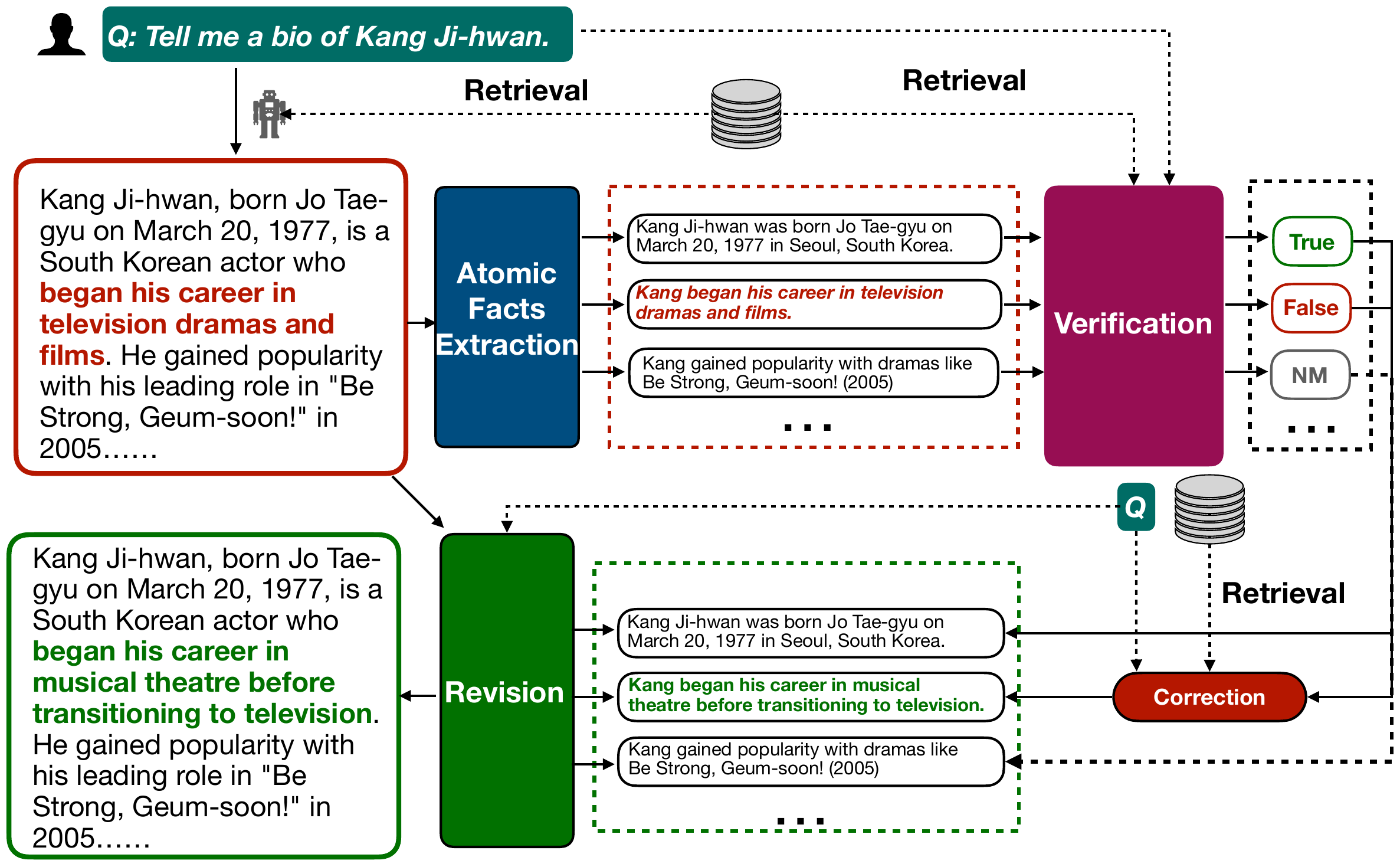}
\caption{Approach overview with RAG. NM means fact not mentioned in the retrieved documents. For LLM outputs with RAG, most content is correct; we only need to correct false ones.}
\label{fig:overview2}
\end{figure*}

\subsection{Overview}
Figures~\ref{fig:overview} and \ref{fig:overview2} show the approach overview with or without RAG. We first break down all original outputs from LLMs into atomic facts \cite{min2023factscore}, which allows our method to do a fine-grained correction of individual facts.

For LLMs without RAG, we propose to add correction and revision stages. The correction stage directly corrects the extracted incorrect statements and keeps the correct statements based on the retrieved document sets for the task input. The statements are then fed into the revision module to revise the original task outputs from LLMs.

For LLMs with RAG, correction by itself is not sufficient. We find that RAG by itself performs well enough that many of the statements do not need to be corrected; simply correcting all statements will introduce more hallucinations, which harm the performance rather than benefit. Considering this, we add a verification component to first to verify the statements and then only correct the false statements. This reduces the hallucinations during RAC since it ensures that truth statements are kept without passing into the correction stage.
 
Both our proposed correction and verification stages use a retriever. The retriever retrieves related factual documents from the input. We apply post-processing for the retrieved documents to make them related, faithful, and concise.

The paper is organized as follows. We introduce our retrieval method and post-processing (\S\ref{Retrieval}), describe atomic statement extraction (\S\ref{facts-split}), and introduce the correction (\S\ref{rac}) and revision (\S\ref{revision}) stages for LLMs without RAG. We review the basics of RAG (\S\ref{rag}) and discuss the additional verification stage (\S\ref{rav}), and the corresponding correction (\S\ref{rac-after-rav}) and revision (\S\ref{revision-after-rav}) stages when used with RAG.  Finally, we present experiments (\S\ref{sec:experiments}), results (\S\ref{sec:results}), ablations (\S\ref{sec:ablations}), and a case study (\S\ref{sec:study}).
 

\subsection{Retrieval}
\label{Retrieval}
The retrieval step includes two parts: retrieval and retrieval post-processing. Retrieval directly retrieves the factual documents using task input from a trusted knowledge source \cite{guu2020realm,NEURIPS2020_6b493230,izacard2022atlas}. The retrieval post-processing conducts two things: 1. filtering out unrelated documents or reranking and picking up top-k documents; 2. truncating or compressing the documents based on the LLM maximum context window length.

Let $X$ be the task input and $R$ be the retrieved outputs.  The retrieval and retrieval post-processing are formulated as follows:
\begin{equation}
    R = \mathrm{\texttt{Retrieve}}(X) 
\end{equation}
\begin{equation}
    R' = \mathrm{\texttt{Compress}}(\mathrm{\texttt{Rerank}}(R))
\end{equation}
$R'$ represents the post-processed retrieved documents.  Our post-processing contains two operations: $\mathrm{\texttt{Compress}}$, which truncates and compresses, and $\mathrm{\texttt{Rerank}}$, which filters and reranks the documents.

The retrieval post-processing ensures the retrieved documents are related, faithful, and compact for the task input. This is important since incorrect or unfaithful retrieved contents can directly cause the failure of the proposed approach, as intuitively, one cannot use incorrect contents to verify and correct a statement. In the ablation studies, we will discuss how the correctness of the retrieved content can affect our approach. The details of retrieval and post-processing processes are in the appendix.

\subsection{Atomic Fact Extraction}
\label{facts-split}
Atomic fact extraction breaks down the original task outputs from LLMs into several independent factual statements. This strategy is inspired by Factscore \cite{min2023factscore}. This can lead to more detailed corrections to the statements, further enhancing the correction and providing a clear interpretation of which part of the original outputs are corrected.

Let the LLM task outputs be $M_{out}$, and the extracted atomic facts $S$. Then:
\begin{equation}
    S = \mathrm{\texttt{Extract}}(M_{out}) = \{s_1, s_2, s_3, ..., s_n\}
\end{equation}
where $n$ is the number of the statements, $s_i$ is the $i$th atomic fact.

\subsection{Correction (C)}
\label{rac}
We add a retrieval process into the factual correction stage, which improves upon the self-correction of prior work~\cite{wang2024finegrained}. When the retrieval document set is trustworthy, this step enhances factual correction. The correction stage corrects the statements based on the verification results using retrieved document sets; then, the revision stage can use them to revise the original LLM task output. This stage ensures that all statements fed into the revision stage are faithful. We also include the task input $X$ during correction to avoid diverging from the task input. The post-processed statements $C$ are:
\begin{equation}
\begin{aligned}
    C = \{\mathrm{\texttt{Correct}}(s_1, R', X), ..., \\
    \mathrm{\texttt{Correct}}(s_n, R', X)\}
\end{aligned}
\end{equation} 

\subsection{Revision (\textit{R})}
\label{revision}
While most works of self-revision use LLMs generated feedback, which can hallucinate all the time \cite{madaan2023selfrefine, selfee2023},  we applied the previous corrected statements after correction during the revision stage, which enables the generated content to align with the ensured truths instead of the hallucinated feedback. The revision stage uses previously corrected statements to revise the original task outputs. To enable the revision to be still consistent with the task input, we include the task input $X$, and the revised outputs $O$ is:
\begin{equation}
    O = \mathrm{\texttt{Revise}}(X, M_{out}, C)
\end{equation}

\section{Combining RAC with RAG}

To further improve results, we can combine our proposed method with retrieval augmented generation (RAG), which we discuss in this section.

\subsection{RAG for LLMs}
\label{rag}
We first review retrieval augmented generation for LLMs.

Given an input $X$, RAG first retrieves documents related to $X$ from a document set $D = \{d_1,....d_m\}$ ($d_i$ represents the $i$th document) to obtain a relevant document set $R = \{d_1,..., d_n\}$. The generation probability $Y$ is the standard next-token prediction probability conditioned on the input context $X$ and retrieved relevant documents $R$.
\begin{equation}
   P(Y|X, R)
\end{equation}

\jmf{Need to say we have found that retrieved content is very important...} \jmf{We need to say we use the top k results from Google, and that it vastly improves over previous approaches. To make this claim, we need to compare to previous approaches to retrieve.} Assuming the retrieved documents are all related to the task input and contain the correct answers, RAG still suffers from potential hallucination issues because retrieved documents may contain other unrelated information that could cause hallucination \cite{pmlr-v202-shi23a}, and the retrieved document may contradict what the model initially learns internally and the model sticks to their original training because of internal prior is very strong during training \cite{wu2024faithful}. To alleviate the above issues, we need to take extra steps to verify and correct using retrieved content during post-generation steps to further reduce the hallucination caused by the above reasons.

\subsection{Verification (V)}
\label{rav}
\jmf{need to explain why we add this for the RAG approach} \cm{done}
\jmf{expand this}Since many of the extracted statements after using RAG are correct, we find we do not need to correct all statements. Instead, we added a verification stage to enable the LLMs to correct only false statements, which reduces the hallucinations introduced by correcting already correct statements.  Some previous self-verification works consider only LLM self-consistency \cite{manakul-etal-2023-selfcheckgpt} or require additional models for verification \cite{mündler2024selfcontradictory}. Unlike these previous works, we add a retrieval process to the verification. The verification is done using LLMs without additional training. The verification stage verifies the extracted atomic facts using the retrieved documents. The verified results are then fed into the correction stage.

Let $V$ be the verification results, Then:
\begin{equation}
    V =  \mathrm{\texttt{Verify}}(R', S) = \{b_1, b_2, b_3, ..., b_n\}
\end{equation}
where $b_i$ is the verification result for the $i$th atomic fact. The value of $b_i$ is \textbf{True}, or \textbf{False}, or \textbf{Not Mentioned}.  \textbf{True} means a similar statement can be found in the retrieved documents and has the same meaning, which indicates the statement is consistent with the retrieved documents. \textbf{False} means a similar statement can be found in the retrieved documents but has a different meaning, which indicates the statement contradicts the retrieved documents. \textbf{Not Mentioned} means a similar statement cannot be found in the retrieved documents, which indicates the statement cannot be verified by the retrieved documents.\jmf{compare our approach to prior work} \cm{Don}

\subsection{Correction (C)}
\label{rac-after-rav}
Let $S_t$, $S_f$, and $S_{nm}$ be the set of atomic facts labeled by the verifier as True, False, or Not Mentioned, respectively.
We use the following strategy to make the correction:
\begin{equation}
C = S_t \cup \{\mathrm{\texttt{Correct}}(s, R', X) |s \in S_f\} \cup S_{nm}
\end{equation}
where True statements are always kept, False statements always be corrected, and Not Mentioned statements are kept.  We also experimented with a strategy that removes all not mentioned statements, but in preliminary experiments found it to give worse results:
\begin{equation}
C = S_t \cup \{\mathrm{\texttt{Correct}}(s, R', X) |s \in S_f\}
\end{equation}

\subsection{Revision (R)}
\label{revision-after-rav}
The revision stage with the verification stage is the same as the revision stage without verification (see \S\ref{revision}). However, to avoid more newly introduced hallucinations for the initial model generations during revision, we also tried a Keep All True (KAT) strategy: only revise model generations with one or more incorrect statements during verification and keep those without any incorrect statements unchanged. Our ablation study in the appendix (\S\ref{sec:kat_ablation}) analyzes the performance of this strategy.

\begin{table*}[t!]
\centering\resizebox{\textwidth}{!}{
\begin{tabular}{l|l|l|l|l|l|l|l|l}
& \multicolumn{2}{c|}{\textbf{Biography}} & \multicolumn{6}{c}{\textbf{TruthfulQA}} \\\hline
   & \multicolumn{2}{c|}{Factscore} & BLEURT      & BLEU        & ROUGE & BLEURT     & BLEU        & ROUGE     \\  \hline
    & \textbf{w/o RAG} & \textbf{w/ RAG}  &\multicolumn{3}{c|}{\textbf{w/o RAG}} & \multicolumn{3}{c}{\textbf{w/ RAG}}\\ \hline \hline 
\multicolumn{9}{c}{\textbf{Llama2-7B} (With Additional Training)}        \\ \hline \hline
  \textit{DoLA} \cite{chuang2024dola}   &39.0  & -  & 39.0  & 36.4 & 36.0  & -  & -  & - \\ 
  \textit{FACTTUNE-MC} \cite{tian2024finetuning} & 42.7 & -  & -  & - & -  & -  & -  & - \\ 
  \textit{SELF-EVAL-SKT} \cite{zhang2024selfalignment}  & 46.5 & -  & -  & - & -  & -  & -  & - \\
 \textit{Self-RAG$^*$(\cite{asai2023selfrag})}     & - &     81.2                      & -  & - & -  & 52.8 & 41.2 & 43.9 \\   
  \textit{Self-CRAG$^*$(\cite{yan2024corrective})}           & - &     86.2                    & -  & - & -  & 53.4 & 40.0 & 40.1 \\    \hline \hline  
 \multicolumn{9}{c}{\textbf{Instruct LLMs} (Without Additional Training)}        \\ \hline \hline
\textbf{\textit{GPT-3.5-Turbo}}                 &78.9 & 93.4              & 58.3  & 48.7 & 51.5  & 67.2  & 62.2& 65.0\\
+ \textit{RARR$^*$ \cite{gao-etal-2023-rarr}}        &79.6 &  85.5      & 58.9 & 47.4  & 50.6 & 63.3  & 58.1  & 62.3 \\
+ \textit{CRITIC$^*$ \cite{gou2024critic} } & 85.2  &  92.0       & 60.8  & 47.2  & 52.0  & 61.8  & 46.3  & 51.0\\
+ \textit{EVER$^*$ \cite{kang2024evermitigatinghallucinationlarge}} & \textbf{93.3}  &  \textbf{93.7}        & 59.6 & 49.2   & 50.6  & 55.7    & 44.6  &  48.8 \\
+ \textit{RAC (ours)}           & 88.2 &  \textbf{93.7} &   \textbf{61.9} & \textbf{50.7} & \textbf{52.8} & \textbf{70.4} & \textbf{67.9} & \textbf{70.4} \\  \hline
\textbf{\textit{Llama2-7B-Chat}}  & 48.7 & 90.2 & 60.7  & 49.9  & 55.2 & 63.4  & 54.8  & 53.6 \\
+ \textit{RAC (ours)}  & \textbf{79.8}  & \textbf{91.5}   & \textbf{77.8}  & \textbf{84.5}  & \textbf{76.5} & \textbf{77.0}  & \textbf{80.2}  & \textbf{72.0}\\ \hline

\textbf{\textit{LLama3-8B-Instruct}}  & 51.1 & 91.0  & 60.7  & 49.9  & 55.2  & 61.1  & 52.1 & 53.4\\
+ \textit{RAC (ours)}  & \textbf{82.6} &  \textbf{92.1}  & \textbf{70.8}                &     \textbf{62.3}           & \textbf{65.0} & \textbf{65.5}                &     \textbf{57.5}           & \textbf{61.9}\\ \hline 
\textbf{\textit{Mistral-7B-Instruct}}  & 49.8 & 90.3  & 67.7  & 54.6  & 56.8  & 63.2  & 51.2 & 51.3\\
+ \textit{RAC (ours)}  & \textbf{80.0} & \textbf{91.2}   & \textbf{67.9}                &     \textbf{59.0}           & \textbf{62.3} & \textbf{65.0}                &     \textbf{53.6}           & \textbf{55.2}\\
\end{tabular}}
\caption{Experimental results on Biography and TruthfulQA. BLEURT, BLEU, and ROUGE are accuracy scores (see \S\ref{datasets}). We report numbers with retrieval augmented generation (RAG) and without RAG. * indicates we reproduced a previous approach using the same retrieved documents and LLM as our approach for a fair comparison. \jmf{remove lower performance model, say in text which one we use. move full results (ablation) to appendix}\jmf{add results from Asai 2023 (and/or run with our RAG content}}
\label{tab:TruthfulQAresults}
\end{table*}

\section{Experimental Settings}
\label{sec:experiments}

\subsection{Datasets and Metrics}
\label{datasets}

We use the two available datasets for factuality evaluation on open-ended generation (not classification): TruthfulQA~\cite{lin2022TruthfulQA} and biography generation \cite{min2023factscore}. 
\jmf{TODO: add  link back in}
For TruthfulQA we use the generation task, which is a short-form generation task. Following the TruthfulQA evaluation, we report the accuracy of BLEURT \cite{sellam-etal-2020-bleurt}, BLEU \cite{papineni-etal-2002-bleu}, and ROUGE \cite{lin-2004-rouge}. Accuracy is computed by comparing the predictions with correct and incorrect answers collected.\footnote{GPT-Judge accuracy for this task is unavailable because OpenAI deprecated the related evaluation model.} Biography is a long-form generation task where the evaluation metric is Factscore. Factscore uses OpenAI GPT-3 to judge the accuracy of factuality compared to the corresponding Wikipedia biography.  Since GPT-3 is no longer available, all reported numbers for Factscore use GPT-3.5-Turbo-Instruct.
 
\subsection{Models and Baselines}
\label{sec:baselines}
We use \model{GPT-3.5-Turbo} \cite{openai2024}, \model{Llama 2-7B-Chat} \cite{touvron2023Llama}, \model{Llama3-8B-Instruct} \cite{meta2024}, and \model{Mistral-7B-Instruct} \cite{jiang2023mistral} as baseline models to evaluate our method on closed and open LLMs. Please see Appendix~\ref{Hyperparameters} for hyperparameters. 

We report numbers for previous state-of-the-art baselines.  To compare our method to the previous method RARR~\cite{gao-etal-2023-rarr}, CRITIC~\cite{gou2024critic} and EVER~\cite{kang2024evermitigatinghallucinationlarge}, we run them using the same model (\model{GPT-3.5-Turbo}) and search engine (Google search) or retrieved documents as ours. EVER is reproduced in a post-correction manner per sentence rather than correction of each sentence during generation to speed up experiments.

\section{Results}
\label{sec:results}
Results are shown in Table \ref{tab:TruthfulQAresults}. We report our findings for each dataset below.

\paragraph{Results on TruthfulQA}
For the TruthfulQA dataset, our method improves upon all previous methods across all LLMs and metrics with and without RAG.

We note the instruction-tuned model Llama2-7B-Chat is better than previous methods using the Llama2-7B model (models listed under ``Llama2-7B With Additional Training'' in Table \ref{tab:TruthfulQAresults}), in both RAG and non-RAG settings. In RAG settings, previous methods RARR \cite{gao-etal-2023-rarr}, CRITIC \cite{gou2024critic} and EVER \cite{kang2024evermitigatinghallucinationlarge} have a lower performance than GPT-3.5-Turbo, indicating that these methods introduce new hallucinations when applied in the RAG setting.  In contrast, our method improves upon GPT-3.5-Turbo even in the RAG setting.
Across base LLM models, our method improves upon the baseline instruction tuned model by up to approximately 35\% on BLEU accuracy, 18\% on  BLEURT accuracy, and 21\% on ROUGE accuracy without RAG and up to 15\% on BLEURT accuracy, 26\% on BLEU accuracy and 20\% on ROUGE accuracy with RAG.
Surprisingly, our approach with \model{Llama2-7B-Chat} and \model{LLama3-8B-Instruct} without RAG is better with RAG, which indicates there are cases where using our approach without RAG is even better than with RAG.



\begin{table*}[t!]
\centering\resizebox{\textwidth}{!}{
\begin{tabular}{l||l|l||l|l|l||l|l|l}
   & \multicolumn{2}{c||}{\textbf{Biography}} & \multicolumn{6}{c}{\textbf{TruthfulQA}} \\\hline
   & \multicolumn{2}{c||}{Factscore} & BLEURT      & BLEU        & ROUGE & BLEURT     & BLEU        & ROUGE     \\  \hline
   & \textbf{Without RAG} & \textbf{With RAG} &\multicolumn{3}{c||}{\textbf{Without RAG}} & \multicolumn{3}{c}{\textbf{With RAG}}\\ \hline \hline
\textbf{\textit{GPT-3.5-Turbo}}        & 78.9  &  93.4                    & 58.3  & 48.7 & 51.5  & 67.2  & 62.2& 65.0\\
+\textit{Self} \textit{C+R}       & \textbf{88.2}  &  93.0           & \textbf{61.9} & \textbf{50.7} & 52.8 & \textbf{72.6} & 62.3 &  58.8 \\ 
+\textit{Self} \textit{V+C+R}  &88.0 & \textbf{93.6} & 59.7 & 49.2 & \textbf{53.5} & 70.4 & \textbf{67.9} & \textbf{70.4}\\ \hline
\textbf{\textit{Llama2-7B-Chat}} &48.7& 90.2 & 60.7  & 49.9  & 55.2 & 63.4  & 54.8  & 53.6 \\
+\textit{Self} \textit{C+R} &\textbf{79.8}& 81.4  & \textbf{77.8}  & \textbf{84.5}  & \textbf{76.5} & 76.0 &78.0 & 72.0\\
+\textit{Self} \textit{V+C+R} &50.4&90.7 & 67.1 & 73.8 & 67.7 & \textbf{77.0}  & \textbf{80.2}  & \textbf{72.0}\\ 
+\textit{GPT} \textit{C+R} & 77.2& 90.2& 71.1                &     61.6            & 64.5 &70.3 & 60.8 & 64.0\\
+\textit{GPT} \textit{V+C+R} &70.1 & \textbf{91.5}& 62.2 & 53.7 & 59.5 & 68.7                &     58.5            & 61.9\\ \hline

\textbf{\textit{LLama3-8B-Instruct}} &51.1& 91.0& 60.7  & 49.9  & 55.2  & 61.1  & 52.1 & 53.4\\
+\textit{Self} \textit{C+R} &76.7& 89.4 & 60.2    & 51.2  &  55.2 &57.9 &49.1 & 52.9\\
+\textit{Self} \textit{V+C+R} &54.8&91.0  & 56.5   & 47.6  & 51.9 & 58.0   &  44.3  & 51.4\\
+\textit{GPT} \textit{C+R} &\textbf{82.6}& 90.5 & \textbf{70.8}                &     \textbf{62.3}           & \textbf{65.0} &\textbf{67.7}&55.6 &58.3\\
+\textit{GPT} \textit{V+C+R} &73.3&\textbf{92.1} &  63.8  & 55.8  & 58.9 & 65.5                &     \textbf{57.5}           & \textbf{61.9}\\ \hline 
\textbf{\textit{Mistral-7B-Instruct}} &49.8& 90.3 & 67.7  & 54.6  & 56.8  & 63.2  & 51.2 & 51.3\\
+\textit{Self} \textit{C+R} &\textbf{80.0}& 90.5 & 64.4   & 54.2 & 55.0 & 60.7   &  51.8  & 52.1\\
+\textit{Self} \textit{V+C+R} &53.0& 90.8  & 64.4    & 51.4  &  53.4 &60.5 &50.1 & 52.4\\
+\textit{GPT} \textit{C+R} &72.2& 89.4  & \textbf{67.9}                &     \textbf{59.0}           & \textbf{62.3} &\textbf{66.3}&\textbf{54.6} &\textbf{56.8}\\
+\textit{GPT} \textit{V+C+R} &68.2& \textbf{91.2} &  67.8  & 55.0  & 60.2 & 65.0                &     53.6           & 55.2\\ 
\end{tabular}}
\caption{Ablation results for Biography and TruthfulQA. Self means the models of \textit{V}, \textit{C} and \textit{R} are the same as the baseline models, \model{GPT} means the models of them are \model{GPT-3.5-turbo} when the baseline is not the \model{GPT-3.5-turbo}. BLEURT, BLEU, and ROUGE are accuracy scores (see \S\ref{datasets}).}
\label{tab:allresultsfull}
\end{table*}

\paragraph{Results on Biography}
For the Biography dataset, our method improves upon all previous methods across all LLMs and metrics with and without RAG, with the exception of our re-implementation of EVER.  However, we note that EVER is much slower than our method (see \S\ref{ssec:latency}).

Similar to TruthfulQA, we note the instruction-tuned model \model{Llama2-7B-Chat} is better than previous methods using \model{Llama2-7B} model, in both RAG and non-RAG settings. The previous approaches RARR and CRITIC improve performance slightly without RAG but have a degraded performance with RAG.  In contrast, our method improves performance by up to 31\% without RAG across three open-sourced models and up to 1.5\% with RAG compared to strong instruction-tuned baselines. Although EVER is slightly better than our method with and without RAG setting, EVER's latency is much larger (see~\S\ref{ssec:latency}).
Considering the baseline RAG performance is already over 90\% in this dataset, our method still shows robust improvement with and without RAG settings across the range of LLMs, especially for open-sourced models. 

\section{Ablation Experiments}
\label{sec:ablations}

\subsection{Ablation of Verification}
\label{sec:abl_verification_stage}
Table~\ref{tab:allresultsfull} shows ablation results with or without verification, and with or without RAG. For LLMs without RAG, in most cases, performance drops significantly after adding the verification, although the performance is still better than the baseline. The reason for this is that without RAG, the original generated content has more content that needs to be corrected, and the verification step removes some critical corrections. For LLMs with RAG, the situation is different and verification improves performance. The reason is that RAG's performance is already very high, so if we correct all the statements, the correction process may introduce hallucinations which lowers the performance.

To conclude, for models without RAG, correcting all statements is optimal, regardless of whether statements are true or false. In the RAG setting, adding a verification stage and correcting only false statements avoids introducing hallucinations during correction and revision.

\subsection{Different LLMs Capabilities}
\label{modelcapability}
Based on the analysis of the above results, we can infer the performance comparison of verification and correction with revision for selected LLMs in different RAG settings for each task. Table \ref{tab:modelabliltiy} in the appendix shows model ability ranking for each component inferred from the results. Generally, \model{Llama2-7B-chat} has the best performance among all settings, while \model{LLama3-8B-Instruct} has the worst performance. While Llama series performance is not stable across the dataset (either \model{Llama2-7B-chat} or \model{LLama3-8B-Instruct} has been ranked third in one or more settings and components), the performance of \model{GPT-3.5-turbo} has not been ranked third, indicating that the closed-source model is more robust than the open-sourced model. The model ability is also task-related, i.e., \model{Mistral-7B-instruct} performs decently in the Biography dataset but poorly in the TruthfulQA.
\begin{table}[htbp!]
\resizebox{\linewidth}{!}{
\begin{tabular}{l|l}
Approach                                       & Factscore \\ \hline
\multicolumn{2}{c}{\textit{GPT3.5-Turbo}}                                     \\ \hline
\textit{C + R}                         & 88.2               \\
\textit{C + R w/ Gold Retrieved Docs}               & 88.4               \\
\textit{RAG w/o filter} &                         93.1                \\
\textit{RAG}                                    & 93.4               \\
\textit{RAG w/o filter +  V+C+R} & 93.5 \\
\textit{RAG + V+C+R} & 93.6               \\
\textit{RAG w/ Gold Retrieved Docs}                         & 97.6               \\
\textit{RAG + V+C+R w/ Gold Retrieved Docs}    & \textbf{97.8}              
\end{tabular}}
\caption{Results comparison of using and without using gold data for the Biography datatset as retrieval documents.\jmf{add ablations of the filtering we use}}
\label{tab:gold-oracle}
\end{table}
\subsection{Effect of Retrieval Correctness}
To analyze the effect of the retrieval correctness, we tested the performance using the gold data from the Biography dataset as the retrieved documents instead of our retrieving methods since this can ensure that the retrieval process is 100\% accurate. We use \model{GPT-3.5-turbo} as the verification, correction, and revision model since it is the most robust model. The results are even promising compared to our sub-optimal retrieving accuracy. Table~\ref{tab:gold-oracle} shows the results of this case. Without RAG, using gold data as the retrieval data for correction only improves the performance little.  RAG using gold data has improved RAG a lot, and our approach can further enhance the RAG with gold retrieved data, achieving a performance of nearly 98\%. This demonstrates that high-quality retrieved data is important to the success of our approach.

\section{Latency}
\label{ssec:latency}

Table~\ref{tab:real-latency} shows the experimentally measured latency on the Biography dataset for our method and previous approaches. Our method has reduced latency of 2x to 40x compared to previous approaches.

We describe the major sources of latency for each method.  RARR generates a set of questions for each sentence in the output and then performs retrieval and reranking for each question, which introduces latency. CRITIC has several correction iterations, increasing the number of LLMs API calls and retrieval calls. EVER generates and corrects the output sentence by sentence, and for each sentence, retrieves using three different types of information; although the performance is slightly better than ours on the Biography dataset without RAG, the latency is the largest of all approaches and may be unacceptable for some applications. In contrast to prior methods, our method retrieves once and corrects once, which reduces latency while remaining highly effective.
\begin{table}[htbp!]
\centering
\begin{tabular}{l|l}
                & Latency \\ \hline
Generation & 1 x                                                                        \\
RARR            & 70 x                                                                          \\
CRITIC          & 10 x                                                                         \\
CRITIC*         & 8.1 x                                                                           \\
EVER            & 150 x                                                                          \\
EVER*           & 98 x                                                                          \\
RAC (ours)       & 3.9 x                                                                          
\end{tabular}
\caption{Experimentally measured latency relative to uncorrected RAG generation for different methods on the Biography dataset. * indicates using the same retrieved documents and LLM as our approach for a fair comparison. EVER* corrects each sentence after generating all sentences of the output, rather than correction of each sentence during generation used in EVER.}
\label{tab:real-latency}
\end{table}

\section{Case Study}
\label{sec:study}

We analyze several examples manually to see the effect of our method. We find the baseline LLM often generates hallucinated content, which is factually incorrect. After applying our correction and revision on this setting without using RAG, all errors are corrected. However, there is still missing information. Using just RAG, the LLMs generate mostly factually correct answers, but there are still some factually incorrect texts. Only applying the correction and revision in this setting may introduce new factual errors since most statements are correct. However, after we add the verification process, correction and revision only get applied to the original texts with errors, which further improves the performance without introducing new factual errors for the original statements that have already been verified as truths. In a manual comparision to RARR \cite{gao-etal-2023-rarr}, we find RARR consistently misses information and introduces more hallucinations, while our method retains facts of the original LLM output and rarely introduce more hallucinations. Table~\ref{tab:case-study} in the Appendix gives an example of generation output.

\section{Conclusion}
We introduce a simple but effective post-processing approach for improving factual correctness for instruction-tuned LLMs. Our method has improved latency over prior methods, does not involve additional training, and can be applied to settings with and without RAG. Experiments demonstrate that the proposed Retrieval Augmented Correction (RAC) approach can enhance the performance on two popular factuality evaluation datasets by up to 30\% for various LLMs and generation setups. For future work, we suggest focusing on reducing newly introduced hallucinations during those operations and improving performance for each operation.

\section*{Limitations}
In rare cases, new hallucinations may be introduced during correction and revision, which should be further investigated in future work. Our verification, correction, and revision prompts for each LLM are not highly optimized but can be tuned for the application. Our approach requires high-quality retrieval data, which may not be available or may require additional steps to acquire it. Like other post-correction methods, our method increased the latency compared to the original generation. Due to budget and hardware constraints constraints, we were not able to experiment with our approach on GPT-4 or on larger open-sourced LLMs.

\section*{Acknowledgements}
We are thankful for the computing resources provided by the Pacific Research Platform's Nautilus cluster, supported in part by National Science Foundation (NSF) awards CNS-1730158, ACI-1540112, ACI-1541349, OAC-1826967, OAC-2112167, CNS-2100237, CNS-2120019, the University of California Office of the President, and the University of California San Diego's California Institute for Telecommunications and Information Technology/Qualcomm Institute, and CENIC for the 100Gbps networks.

\bibliography{anthology,custom}

\appendix

\section{Retrieval \& Post-Processing}
We apply Google search for the retrieval process to obtain high-quality retrieval data. We then applied different post-processing strategies for different tasks since different tasks have different features.

For the biography task, we use the keyword "\{Named Entity\} Wikipedia" to search Google since the biography dataset is mainly from Wikipedia. After retrieving the top 10 results, we have two stages for the postprocessing; one is the filtering, and the other is truncating to fit the input length of LLMs. We first filter out the results that do not have the searched named entity, increasing the retrieval performance to a very good level. We then picked the first result from Wikipedia among the filtered results. Then, we truncate Wikipedia's useless sections for generating biographies, such as "References," "Footnotes," "Notes and references," "Notes," etc. After that, if the length is still too long, we remove some non-textual sections for generating biographies, such as the "Filmography" list, "Production" list, "Career statistics" table, etc. After removing them, most retrieved content is enough to fit the LLMs' context.

For the TruthfulQA task, we use their questions to search Google. Note that we found that Google search was contaminated with the dataset in this case since we can find the exact match of the answers in the dataset. Considering this, we retrieved the top 30 results and removed all of those data-leaking results by links such as "huggingface", "paperswithcode", "kaggle", "openreview", "github", "arxiv", etc to avoid cheating. We then only keep results that have longer than one sentence since some results are empty with just a hooked title. To fit for LLMs context length, in this case, we directly truncate all retrieved content to a fixed length since most of the related answers are on the very first sections of a retrieved page.

\section{Hyperparameters}
\label{Hyperparameters}
For GPT-3.5-turbo, we use nucleus sampling with top\_p = 0.3, meaning only the tokens comprising the top 30\% probability mass are considered during generation. For Llama 2-Chat-7B or Llama 3-Chat-7B, we use their default setting. For different approaches, the hyperparameter settings for each LLM are the same.

\section{Performance Ranking for Each Model}
Table \ref{tab:modelabliltiy} shows performance ranking for each LLM on different proposed operations based on evaluation results.

\begin{table*}[htbp!]
\small\centering
\begin{tabular}{l|l|l}
& \multicolumn{1}{c}{Biograph} & \multicolumn{1}{|c}{TruthfulQA} \\ \hline
\multicolumn{3}{c}{Without RAG}                                          \\ \hline
Correction + Revision & Llama 2 \textgreater Mistral \textgreater GPT \textgreater Llama 3  & Llama 2 \textgreater GPT \textgreater Llama 3=Mistral  \\ \hline
\multicolumn{3}{c}{With RAG}                                             \\ \hline
Correction + Revision & GPT \textgreater Mistral
\textgreater Llama 3 \textgreater Llama 2 & Llama 2 \textgreater GPT \textgreater  Llama 3=Mistral  \\
Verification performance               & GPT \textgreater Llama 2 \textgreater Mistral \textgreater Llama 3 & Llama 2 \textgreater GPT \textgreater Llama 3=Mistral 
\end{tabular}
\caption{Model ability ranking table for each component inferred from the results. Llama 2 represents the Llama2-7B-Chat model, Llama 3 represents the LLama3-8B-Instruct model, and GPT represents GPT-3.5-turbo, Mistral represents the Mistral-7B-intruct.\jmf{fix table}}
\label{tab:modelabliltiy}
\end{table*}

\section{Keep All True (KAT) Ablation}
\label{sec:kat_ablation}
Table~\ref{tab:kat} shows results comparison of using and without using KAT.
\begin{table*}[t!]
\small \centering
\begin{tabular}{l||l||lll}
                                   & Biograph & \multicolumn{3}{c}{TruthfulQA}     \\ 
                         & Factscore         & BLEURT & BLEU & ROUGE \\ \hline
\multicolumn{5}{c}{GPT-3.5-turbo}     \\ \hline
\textit{RAG +  Self-(V+C+R)}       & 93.6     & 70.4       & 67.9     & 70.4       \\
\textit{RAG +  Self-(V+C+R) + KAT} & 93.7     & 68.4       & 64.3     & 66.6       \\ \hline
\multicolumn{5}{c}{Llama2-7B-Chat}     \\ \hline
\textit{RAG +  Self-(V+C+R)}      & 91.5     & 77.0         & 80.2     & 72.0         \\
\textit{RAG +  Self-(V+C+R)+KAT}  & 90.6     & 67.2       & 62.5     & 59.6       \\ \hline
\multicolumn{5}{c}{LLama3-8B-Instruct}     \\ \hline
\textit{RAG + Self-(V+C+R)}       & 92.1     & 58.0         & 44.3     & 51.4       \\
\textit{RAG + Self-(V+C+R)+KAT}   & 91.5     & 59.0         & 49.1     & 52.9      
\end{tabular}
\caption{Keep All True results}
\label{tab:kat}
\end{table*}

\section{Prompts}
Table~\ref{tab:task-prompts} shows prompts for each operation.
\begin{table*}[t!]
\small\centering
\begin{tabular}{p{0.22\linewidth}|p{0.77\linewidth}} 
Operation & Prompt \\\hline
RAG for Biography &
  \{Task Question\}\textbackslash{}nAnswer based on the following text and keep the answer authentic to the texts:\textbackslash{}n\textbackslash{}"\{retrieved documents\}\textbackslash{}"\textbackslash{}n\textbackslash{}nAnswer:\textbackslash{}n \\ \hline
RAG for TruthfulQA &
  Passages:"\{retrieved documents\}"\textbackslash{}n\{Task Question\}\textbackslash{}nPlease find the answer to the question from the above passages and generate the answer text. If there is an answer in the documents, please keep the answer authentic to the passage, if the question is to ask for opinion or if there is no answer found in the documents, please output "I have no comment".\textbackslash{}nAnswer:\textbackslash{}n") \\ \hline
Atomic Fact Extraction &
  Please breakdown the following content into independent facts without pronouns(Do not use He, She, It...)(each fact should be a full sentence, each fact per line):"origiinal model generation"\textbackslash{}nFacts:\textbackslash{}n \\ \hline
Verification &
  \{Task Question\}\textbackslash{}npassage:"\{retrieved documents\}\textbackslash{}"\textbackslash{}nPlease verify the below statements to the above question into true or false or not mentioned based on the above passages (one answer per line with label true or false or not mentioned.)\textbackslash{}nTrue means the similar statement can be found in the above passage and have the same meaning.\textbackslash{}nFalse means the similar statement can be found in the above passage  but have the different meaning.\textbackslash{}nNot Mentioned means the similar statement cannot be found in the above passage.\textbackslash{}n\textbackslash{}nStatements:"\{extracted atomic facts\}"\textbackslash{}n\textbackslash{}nOutput Format:\textbackslash{}nStatement 1: True\textbackslash{}nStatement 2: False \textbackslash{}n ... \textbackslash{}nStatement N: Not Mentioned\textbackslash{}n\textbackslash{}nAnswer(start with the output directly without additional comments):\textbackslash{}n") \\ \hline
Correction for all &
  \{Task Question\}\textbackslash{}npassage:"\{retrieved documents\}"\textbackslash{}nCorrect the following statement and output the corrected version based on the above passage. If the statement is correct, directly output the original statement. In your answer, start with the corrected answer or original correct statement directly without repeating the question. The answer should be a single sentence and should be concise and to the point of the question. \textbackslash{}n\textbackslash{}nStatement:"\{extracted atomic facts\}"\textbackslash{}n\textbackslash{}nAnswer:\textbackslash{}n \\ \hline
Correction for False &
  \{Task Question\}\textbackslash{}npassage:"\{retrieved documents\}"\textbackslash{}nCorrect the following statement and output the corrected version based on the above passage. In your answer, start with the corrected answer directly without repeating the question or the original statement. \textbackslash{}n\textbackslash{}nStatement:\textbackslash{}"\{a false statement after RAV\}\textbackslash{}"\textbackslash{}n\textbackslash{}nAnswer:\textbackslash{}n" \\ \hline
Revision for Biography &
  \{Task Question and Original Generated Answer\}\textbackslash{}n\textbackslash{}nPlease correct the above answer into a corrected one based on the following verified facts. In your answer, start with the corrected answer directly without repeating the question or the original answer.\textbackslash{}n\textbackslash{}nVerified facts:"\{statements after correction\}"\textbackslash{}n\textbackslash{}nCorrected answer:\textbackslash{}n \\ \hline
Revision for TruthfulQA &
  \{Task Question and Original Generated Answer\}\textbackslash{}n\textbackslash{}nPlease correct the above answer into a corrected one based on the following verified facts. In your answer, start with the corrected answer directly without repeating the question or the original answer. if the answer is "I have no comment", output "I have no comment".\textbackslash{}n\textbackslash{}nVerified facts:"\{statements after correction\}"\textbackslash{}n\textbackslash{}nCorrected answer:\textbackslash{}n
\end{tabular}
\caption{Prompts for each operation}
\label{tab:task-prompts}
\end{table*}

\section{Example Output}
Table~\ref{tab:case-study} shows the outputs for each stage for an example from the Biography dataset.

\begin{table*}[t!]
\tiny\centering
\begin{tabular}{p{0.16\linewidth}|p{0.80\linewidth}}
Component &
  Task: Biography \\ \hline
Llama2-7B-Chat &
 Sara Paxton is an American actress, singer\textbf{, and model. She was born on November 25, 1988, in Woodridge, New Jersey,} USA. Her parents are Lucille and Steve Paxton. \textbf{She has one older brother, Sean Paxton.} Sara began her acting career at a young age, appearing in various TV commercials and films. \textbf{Her breakthrough role came in 2003 when she played the lead role in the Disney Channel original movie "Summerland."} Her performance earned her a Daytime Emmy Award nomination for Outstanding Young Actress in a Drama Series. \textbf{In 2005, Sara landed a recurring role on the hit TV series "Dirt," playing the character of Riley Richmond.} She also appeared in the films "Aquamarine" and "The Last House on the Left." \textbf{In 2009, Sara starred in the horror movie "The Last Exorcism," which was a box office success. She also released her debut album, "Sara Paxton," which included the hit single "Birthday."} Sara has continued to act and sing in various projects, \textbf{including the TV series "Houdini \& Doyle" and the film "The Circle."} \textbf{She has also been a guest judge on the reality TV show "American Idol."} Sara is known for her versatility and range as an actress and singer. She has been praised for her performances in various genres, from comedy to drama to horror. \textbf{She is also recognized for her philanthropic work, particularly in the area of animal welfare.} \\ \hline
RARR \cite{gao-etal-2023-rarr} & Sara Paxton is an American actress and singer, born on April 25, 1988, in Woodland Hills, California. She began her acting career at a young age, appearing in various television shows and commercials. Paxton gained recognition for her role as Mari Collingwood in the horror film "The Last House on the Left" (2009) and \textbf{as Amanda in the teen comedy film "Aquamarine" (2006)}.

In addition to her acting career, Paxton is also a talented singer and has released several singles and music videos. She has also lent her voice to various animated projects, \textbf{including the role of Marnie in the "Happily N'Ever After" film series}.

Paxton continues to work in film and television, with recent credits including roles in the TV series "Good Girls" and the film "The Front Runner" (2018). She is known for her versatility as an actress and her ability to portray a wide range of characters across different genres. \\ \hline

GPT (C+R) &
Sara Paxton is an American actress and singer. She was born in Woodland Hills, Los Angeles, California, USA. Her parents are Lucia and Steve Paxton. Sara is the only child of her parents and does not have an older brother named Sean Paxton. Sara began her acting career at an early age, appearing in minor roles in both films and television shows before rising to fame in Early October 2002. Her breakthrough role came in 2004 when she played the role of Sarah Borden in the television series "Summerland." She received an Emmy nomination for Outstanding Young Actress in a Drama Series for her performance in the television series "Darcy's Wild Life." Sara appeared in the films "Aquamarine" and "The Last House on the Left." She did not release a debut album titled "Sara Paxton" with the hit single "Birthday." She has not appeared in the TV series "Houdini \& Doyle" or the film "The Circle." Sara is known for her versatility and range as an actress and singer. She has been praised for her performances in various genres, from comedy to drama to horror. Sara is married to Zach Cregger and resides in Los Angeles, California. Her advocacy work and involvement with animal rights organizations to promote animal welfare are not mentioned in the passage. In her free time, she enjoys singing, writing music, and spending time with her pets. Sara received an Emmy nomination for her role in the Discovery Kids television series, Darcy's Wild Life, in 2006, not in 2003. Her breakthrough role was in the television series "Summerland" in 2004, not in the Disney Channel original movie "Summerland" in 2003. Sara did not have a starring role in the horror movie "The Last Exorcism" in 2009. She did not release her debut album in 2009. \\ \hline
RAG & Sara Paxton is an American actress and singer \textbf{who was born on November 25, 1988}, in Woodland Hills, Los Angeles, California. She began her acting career at a young age, appearing in minor roles in both films and television shows before rising to fame in 2002. Paxton has starred in numerous films and television series, including "Aquamarine," "Return to Halloweentown," "Sydney White," "Superhero Movie," "The Last House on the Left," and "The Innkeepers." She has also provided backing vocals on her co-star Drake Bell's theme song for the movie "Superhero!" Song. Paxton was raised in the San Fernando Valley and graduated from El Camino Real High School in 2006. She did not attend college, choosing instead to pursue her acting career. Paxton has been nominated for several awards, including an Emmy nomination for her role in the Discovery Kids television series "Darcy's Wild Life." In her personal life, Paxton married Zach Cregger in October 2019. \\ \hline
RAG Atomic Facts &
  \begin{tabular}[]{@{}p{\linewidth}@{}}Sara Paxton is an American actress and singer.\\\textbf{ She was born on November 25, 1988,} in Woodland Hills, Los Angeles, California.\\ Paxton began her acting career at a young age.\\ She appeared in minor roles in both films and television shows before rising to fame in 2002.\\ Paxton has starred in numerous films and television series.\\ Some of her notable works include "Aquamarine," "Return to Halloweentown," and "The Last House on the Left."\\ She also provided backing vocals on her co-star Drake Bell's theme song for the movie "Superhero!"\\ Paxton was raised in the San Fernando Valley.\\ She graduated from El Camino Real High School in 2006.\\ Paxton did not attend college, choosing instead to pursue her acting career.\\ She has been nominated for several awards, including an Emmy nomination for her role in the Discovery Kids television series "Darcy's Wild Life."\\ In her personal life, Paxton married Zach Cregger in October 2019.\end{tabular} \\ \hline
RAG verified results &
  Statement 1: True \textbackslash n \textbf{Statement 2: False} \textbackslash n Statement 3: True\textbackslash n Statement 4: True\textbackslash n Statement 5: True\textbackslash n Statement 6: True\textbackslash n Statement 7: True\textbackslash n Statement 8: True\textbackslash n Statement 9: True\textbackslash n Statement 10: True\textbackslash n Statement 11: True\textbackslash n Statement 12: True\\ \hline
RAG + GPT-(V+C+R) &
  Sara Paxton is an American actress and singer. She was born in Woodland Hills, Los Angeles, California. Paxton began her acting career at a young age, appearing in minor roles in both films and television shows before rising to fame in 2002. She has starred in numerous films and television series, including "Aquamarine," "Return to Halloweentown," and "The Last House on the Left." Paxton has also provided backing vocals on her co-star Drake Bell's theme song for the movie "Superhero!" She was raised in the San Fernando Valley and graduated from El Camino Real High School in 2006. Paxton did not attend college, choosing instead to pursue her acting career. She has been nominated for several awards, including an Emmy nomination for her role in the Discovery Kids television series "Darcy's Wild Life." In her personal life, Paxton married Zach Cregger in October 2019.
\end{tabular}
\caption{An example generation output flow from biography using different operations. \textbf{Bold texts} represent incorrect facts. No \textbf{Bold texts} means no factual error. Outputs are formatted to fit the table without changing the original content.}
\label{tab:case-study}
\end{table*}

\end{document}